\documentclass[english]{elsarticle}
\usepackage[T1]{fontenc}
\usepackage[latin9]{inputenc}
\usepackage{array}
\usepackage{float}
\usepackage{multirow}
\usepackage{amsmath}
\usepackage{amsthm}
\usepackage{amssymb}
\usepackage{graphicx}
\usepackage{xargs}[2008/03/08]

\makeatletter

\providecommand{\tabularnewline}{\\}
\floatstyle{ruled}
\newfloat{algorithm}{tbp}{loa}
\providecommand{\algorithmname}{Algorithm}
\floatname{algorithm}{\protect\algorithmname}

\theoremstyle{plain}
\newtheorem{thm}{\protect\theoremname}
  \theoremstyle{plain}
  \newtheorem{lem}[thm]{\protect\lemmaname}

\@ifundefined{date}{}{\date{}}
\usepackage{numcompress}
\usepackage{algolyx}
\journal{xxx}

\makeatother

\usepackage{babel}
  \providecommand{\lemmaname}{Lemma}
\providecommand{\theoremname}{Theorem}

\begin{document}

\begin{frontmatter}{}

\title{Scalable Support Vector Clustering Using Budget}

\author[firstaddress]{Tung Pham}
\author[secondaddress]{Trung Le \corref{correspondingauthor}}
\cortext[correspondingauthor]{Corresponding author}
\ead{trung.l@deakin.edu.au}
\author[firstaddress]{Hang Dang}
\address[firstaddress]{Faculty of Information Technology, VNUHCM-University of Science, Vietnam} 
\address[secondaddress]{Centre for Pattern Recognition and Data Analytics, Deakin University, Australia}
\begin{abstract}
Owing to its application in solving the difficult and diverse clustering
or outlier detection problem, support-based clustering has recently
drawn plenty of attention. Support-based clustering method always
undergoes two phases: finding the domain of novelty and performing
clustering assignment. To find the domain of novelty, the training
time given by the current solvers is typically over-quadratic in the
training size, and hence precluding the usage of support-based clustering
method for large-scale datasets. In this paper, we propose applying
Stochastic Gradient Descent (SGD) framework to the first phase of
support-based clustering for finding the domain of novelty and a new
strategy to perform the clustering assignment. However, the direct
application of SGD to the first phase of support-based clustering
is vulnerable to the curse of kernelization, that is, the model size
linearly grows up with the data size accumulated overtime. To address
this issue, we invoke the budget approach which allows us to restrict
the model size to a small budget. Our new strategy for clustering
assignment enables a fast computation by means of reducing the task
of clustering assignment on the full training set to the same task
on a significantly smaller set. We also provide a rigorous theoretical
analysis about the convergence rate for the proposed method. Finally,
we validate our proposed method on the well-known datasets for clustering
to show that the proposed method offers a comparable clustering quality
while simultaneously achieving significant speedup in comparison with
the baselines.
\end{abstract}

\end{frontmatter}{}

\newcommand{\sidenote}[1]{\marginpar{\small \emph{\color{Medium}#1}}}

\global\long\def\se{\hat{\text{se}}}

\global\long\def\interior{\text{int}}

\global\long\def\boundary{\text{bd}}

\global\long\def\ML{\textsf{ML}}

\global\long\def\GML{\mathsf{GML}}

\global\long\def\HMM{\mathsf{HMM}}

\global\long\def\support{\text{supp}}

\global\long\def\new{\text{*}}

\global\long\def\stir{\text{Stirl}}

\global\long\def\mA{\mathcal{A}}

\global\long\def\mB{\mathcal{B}}

\global\long\def\mF{\mathcal{F}}

\global\long\def\mK{\mathcal{K}}

\global\long\def\mH{\mathcal{H}}

\global\long\def\mX{\mathcal{X}}

\global\long\def\mZ{\mathcal{Z}}

\global\long\def\mS{\mathcal{S}}

\global\long\def\Ical{\mathcal{I}}

\global\long\def\mT{\mathcal{T}}

\global\long\def\Pcal{\mathcal{P}}

\global\long\def\dist{d}

\global\long\def\HX{\entro\left(X\right)}
 \global\long\def\entropyX{\HX}

\global\long\def\HY{\entro\left(Y\right)}
 \global\long\def\entropyY{\HY}

\global\long\def\HXY{\entro\left(X,Y\right)}
 \global\long\def\entropyXY{\HXY}

\global\long\def\mutualXY{\mutual\left(X;Y\right)}
 \global\long\def\mutinfoXY{\mutualXY}

\global\long\def\given{\mid}

\global\long\def\gv{\given}

\global\long\def\goto{\rightarrow}

\global\long\def\asgoto{\stackrel{a.s.}{\longrightarrow}}

\global\long\def\pgoto{\stackrel{p}{\longrightarrow}}

\global\long\def\dgoto{\stackrel{d}{\longrightarrow}}

\global\long\def\lik{\mathcal{L}}

\global\long\def\logll{\mathit{l}}

\global\long\def\vectorize#1{\mathbf{#1}}

\global\long\def\vt#1{\mathbf{#1}}

\global\long\def\gvt#1{\boldsymbol{#1}}

\global\long\def\idp{\ \bot\negthickspace\negthickspace\bot\ }
 \global\long\def\cdp{\idp}

\global\long\def\das{}

\global\long\def\id{\mathbb{I}}

\global\long\def\idarg#1#2{\id\left\{  #1,#2\right\}  }

\global\long\def\iid{\stackrel{\text{iid}}{\sim}}

\global\long\def\bzero{\vt 0}

\global\long\def\bone{\mathbf{1}}

\global\long\def\boldm{\boldsymbol{m}}

\global\long\def\bff{\vt f}

\global\long\def\bx{\boldsymbol{x}}

\global\long\def\bl{\boldsymbol{l}}

\global\long\def\bu{\boldsymbol{u}}

\global\long\def\bo{\boldsymbol{o}}

\global\long\def\bh{\boldsymbol{h}}

\global\long\def\bs{\boldsymbol{s}}

\global\long\def\bz{\boldsymbol{z}}

\global\long\def\xnew{y}

\global\long\def\bxnew{\boldsymbol{y}}

\global\long\def\bX{\boldsymbol{X}}

\global\long\def\tbx{\tilde{\bx}}

\global\long\def\by{\boldsymbol{y}}

\global\long\def\bY{\boldsymbol{Y}}

\global\long\def\bZ{\boldsymbol{Z}}

\global\long\def\bU{\boldsymbol{U}}

\global\long\def\bv{\boldsymbol{v}}

\global\long\def\bn{\boldsymbol{n}}

\global\long\def\bV{\boldsymbol{V}}

\global\long\def\bI{\boldsymbol{I}}

\global\long\def\bw{\vt w}

\global\long\def\balpha{\gvt{\alpha}}

\global\long\def\bbeta{\gvt{\beta}}

\global\long\def\bmu{\gvt{\mu}}

\global\long\def\btheta{\boldsymbol{\theta}}

\global\long\def\blambda{\boldsymbol{\lambda}}

\global\long\def\bgamma{\boldsymbol{\gamma}}

\global\long\def\bpsi{\boldsymbol{\psi}}

\global\long\def\bphi{\boldsymbol{\phi}}

\global\long\def\bpi{\boldsymbol{\pi}}

\global\long\def\bomega{\boldsymbol{\omega}}

\global\long\def\bepsilon{\boldsymbol{\epsilon}}

\global\long\def\btau{\boldsymbol{\tau}}

\global\long\def\bxi{\boldsymbol{\xi}}

\global\long\def\realset{\mathbb{R}}

\global\long\def\realn{\realset^{n}}

\global\long\def\integerset{\mathbb{Z}}

\global\long\def\natset{\integerset}

\global\long\def\integer{\integerset}

\global\long\def\natn{\natset^{n}}

\global\long\def\rational{\mathbb{Q}}

\global\long\def\rationaln{\rational^{n}}

\global\long\def\complexset{\mathbb{C}}

\global\long\def\comp{\complexset}

\global\long\def\compl#1{#1^{\text{c}}}

\global\long\def\and{\cap}

\global\long\def\compn{\comp^{n}}

\global\long\def\comb#1#2{\left({#1\atop #2}\right) }

\global\long\def\nchoosek#1#2{\left({#1\atop #2}\right)}

\global\long\def\param{\vt w}

\global\long\def\Param{\Theta}

\global\long\def\meanparam{\gvt{\mu}}

\global\long\def\Meanparam{\mathcal{M}}

\global\long\def\meanmap{\mathbf{m}}

\global\long\def\logpart{A}

\global\long\def\simplex{\Delta}

\global\long\def\simplexn{\simplex^{n}}

\global\long\def\dirproc{\text{DP}}

\global\long\def\ggproc{\text{GG}}

\global\long\def\DP{\text{DP}}

\global\long\def\ndp{\text{nDP}}

\global\long\def\hdp{\text{HDP}}

\global\long\def\gempdf{\text{GEM}}

\global\long\def\rfs{\text{RFS}}

\global\long\def\bernrfs{\text{BernoulliRFS}}

\global\long\def\poissrfs{\text{PoissonRFS}}

\global\long\def\grad{\gradient}
 \global\long\def\gradient{\nabla}

\global\long\def\partdev#1#2{\partialdev{#1}{#2}}
 \global\long\def\partialdev#1#2{\frac{\partial#1}{\partial#2}}

\global\long\def\partddev#1#2{\partialdevdev{#1}{#2}}
 \global\long\def\partialdevdev#1#2{\frac{\partial^{2}#1}{\partial#2\partial#2^{\top}}}

\global\long\def\closure{\text{cl}}

\global\long\def\cpr#1#2{\Pr\left(#1\ |\ #2\right)}

\global\long\def\var{\text{Var}}

\global\long\def\Var#1{\text{Var}\left[#1\right]}

\global\long\def\cov{\text{Cov}}

\global\long\def\Cov#1{\cov\left[ #1 \right]}

\global\long\def\COV#1#2{\underset{#2}{\cov}\left[ #1 \right]}

\global\long\def\corr{\text{Corr}}

\global\long\def\sst{\text{T}}

\global\long\def\SST{\sst}

\global\long\def\ess{\mathbb{E}}

\global\long\def\Ess#1{\ess\left[#1\right]}

\newcommandx\ESS[2][usedefault, addprefix=\global, 1=]{\underset{#2}{\ess}\left[#1\right]}

\global\long\def\fisher{\mathcal{F}}

\global\long\def\bfield{\mathcal{B}}
 \global\long\def\borel{\mathcal{B}}

\global\long\def\bernpdf{\text{Bernoulli}}

\global\long\def\betapdf{\text{Beta}}

\global\long\def\dirpdf{\text{Dir}}

\global\long\def\gammapdf{\text{Gamma}}

\global\long\def\gaussden#1#2{\text{Normal}\left(#1, #2 \right) }

\global\long\def\gauss{\mathbf{N}}

\global\long\def\gausspdf#1#2#3{\text{Normal}\left( #1 \lcabra{#2, #3}\right) }

\global\long\def\multpdf{\text{Mult}}

\global\long\def\poiss{\text{Pois}}

\global\long\def\poissonpdf{\text{Poisson}}

\global\long\def\pgpdf{\text{PG}}

\global\long\def\wshpdf{\text{Wish}}

\global\long\def\iwshpdf{\text{InvWish}}

\global\long\def\nwpdf{\text{NW}}

\global\long\def\niwpdf{\text{NIW}}

\global\long\def\studentpdf{\text{Student}}

\global\long\def\unipdf{\text{Uni}}

\global\long\def\transp#1{\transpose{#1}}
 \global\long\def\transpose#1{#1^{\mathsf{T}}}

\global\long\def\mgt{\succ}

\global\long\def\mge{\succeq}

\global\long\def\idenmat{\mathbf{I}}

\global\long\def\trace{\mathrm{tr}}

\global\long\def\argmax#1{\underset{_{#1}}{\text{argmax}} }

\global\long\def\argmin#1{\underset{_{#1}}{\text{argmin}\ } }

\global\long\def\diag{\text{diag}}

\global\long\def\norm{}

\global\long\def\spn{\text{span}}

\global\long\def\vtspace{\mathcal{V}}

\global\long\def\field{\mathcal{F}}
 \global\long\def\ffield{\mathcal{F}}

\global\long\def\inner#1#2{\left\langle #1,#2\right\rangle }
 \global\long\def\iprod#1#2{\inner{#1}{#2}}

\global\long\def\dprod#1#2{#1 \cdot#2}

\global\long\def\norm#1{\left\Vert #1\right\Vert }

\global\long\def\entro{\mathbb{H}}

\global\long\def\entropy{\mathbb{H}}

\global\long\def\Entro#1{\entro\left[#1\right]}

\global\long\def\Entropy#1{\Entro{#1}}

\global\long\def\mutinfo{\mathbb{I}}

\global\long\def\relH{\mathit{D}}

\global\long\def\reldiv#1#2{\relH\left(#1||#2\right)}

\global\long\def\KL{KL}

\global\long\def\KLdiv#1#2{\KL\left(#1\parallel#2\right)}
 \global\long\def\KLdivergence#1#2{\KL\left(#1\ \parallel\ #2\right)}

\global\long\def\crossH{\mathcal{C}}
 \global\long\def\crossentropy{\mathcal{C}}

\global\long\def\crossHxy#1#2{\crossentropy\left(#1\parallel#2\right)}

\global\long\def\breg{\text{BD}}

\global\long\def\lcabra#1{\left|#1\right.}

\global\long\def\lbra#1{\lcabra{#1}}

\global\long\def\rcabra#1{\left.#1\right|}

\global\long\def\rbra#1{\rcabra{#1}}

\section{Introduction}

Cluster analysis is a fundamental problem in pattern recognition and
machine learning which has a wide spectrum of applications in reality.
The goal of cluster analysis is to categorize objects into groups
or clusters based on pairwise similarities between those objects such
that two criteria, homogeneity and separation, are achieved \cite{Shamir01algorithmicapproaches}.
 Two challenging tasks in cluster analysis are (1) to deal with complicated
data with nested or hierarchy structures inside; and (2) to automatically
detect the number of clusters. Recently inspired from the seminal
work of \cite{Ben-Hur01supportvector}, support-based clustering has
drawn a significant research concern because of its applications in
solving the difficult and diverse clustering or outlier detection
problem \cite{Ben-Hur01supportvector,yang2002support,ParkJZK04,CamastraV05,JungLL10,Lee2005,LeTNMS13}.
The support-based clustering methods have two main advantages in comparison
to other clustering methods: (1) ability to generate the clustering
boundaries with arbitrary shapes and automatically discover the number
of clusters; and (2) capability to handle well the outliers.

Support-based clustering methods always undergo two phases. In the
first phase (i.e., finding the domain of novelty), an optimal hypersphere
\cite{Ben-Hur01supportvector,tax2004support,le2010optimal,Le_etal_13Fuzzy,Le_etal_13Fuzzya,Le_etal_11Generalised}
or hyperplane \cite{scholkopf2001,Nguyen_etal_14Kernel} is discovered
in the feature space. The domain of novelty when mapped back into
the input space will become a set of contours tightly enclosing data
which can be interpreted as cluster boundaries. However, this set
of contours does not point out the assignment of data samples to clusters.
Furthermore, the computational complexity of the current solvers \cite{joachims1999,libsvm}
to find out the domain of novelty is often over-quadratic \cite{Shalev-ShwartzK08}.
Such a computational complexity renders the application of support-based
clustering methods for the real-world datasets impractical. In the
second phase (i.e., clustering assignment), based on the geometry
information carried in the resultant set of contours obtained from
the first phase, data samples are appointed to their clusters. Several
existing works have been proposed to improve cluster assignment procedure
\cite{yang2002support,ParkJZK04,CamastraV05,Lee2005,JungLL10}. 

Recently, stochastic gradient descent (SGD) frameworks \cite{Robbins&Monro1951,Shalev-shwartz07logarithmicregret,Shalevr07pegasos:primal,HazanK14}
have offered building blocks to develop scalable learning methods
for efficiently handling large-scale dataset. SGD-based algorithm
has the following advantages: (1) very fast; (2) ability to run in
online mode; and (3) economic in memory usage. In this paper, we conjoin
the strengths of SGD with support-based clustering. In particular,
we propose to use the optimal hyperplane as the domain of novelty.
The margin, i.e., the distance from the origin to the optimal hyperplane,
is maximized to make the contours enclosing the data as tightly as
possible. Subsequently, we employ the SGD framework proposed in \cite{Shalev-shwartz07logarithmicregret}
to the first phase of support-based clustering to efficiently solve
the incurred optimization problem. Unfortunately, this direct employment
\cite{Pham2016} is vulnerable to the \emph{curse of kernelization
}\cite{wang2012} wherein the model size linearly grows up with the
data size accumulated over time. To address this issue, we utilize
the budget approach \cite{Crammer04onlineclassification,CavallantiCG07,Wang2010,wang2012,Le2016,TrungUAI2016,Le_etal_16Dual,Le_etal_16Approximation}
to restrict the model size to a budget, and hence keeping the most
important data samples whilst partly preserving the information of
the less important ones. More specifically, a budget maintenance procedure
(e.g., removal, projection, or merging) is invoked to maintain the
model size whenever this model size exceeds the budget. Finally, we
propose a new strategy for clustering assignment where each data sample
in the extended decision boundary has its own trajectory to converge
to an equilibrium point and clustering assignment task is then reduced
to the same task for those equilibrium points. Our clustering assignment
strategy distinguishes from the existing works of \cite{Lee2005,Lee2006,JungLL10,li2013fast}
in the way to find the trajectory with a start and the initial set
of data samples that need to do a trajectory for finding the corresponding
equilibrium point. The experiments established on the real-world datasets
show that our proposed method produces the comparable clustering quality
with other support-based clustering methods while simultaneously achieving
the computational speedup.

To summarize, the contribution of the paper consists of the following
points:
\begin{itemize}
\item In contrast to the works of \cite{Ben-Hur01supportvector,CamastraV05,Lee2005,ParkJZK04,yang2002support}
which employ a hypersphere to describe the domain of novelty, we propose
using a hyperplane to characterize the domain of novelty. This allows
us to introduce SGD-based solution for finding the domain of novelty.
\item We propose budgeted SGD-based solution for finding the domain of novelty
which allows the incurred optimization to be efficiently solved and
to be invulnerable to the curse of kernelization. We perform a rigorous
convergence analysis for the proposed solution. We note that the works
of \cite{Ben-Hur01supportvector,CamastraV05,Lee2005,ParkJZK04,yang2002support}
utilized the Sequential Minimal Optimization based approach \cite{Platt:1999}
to find the domain of novelty wherein the computational complexity
is over-quadratic and the memory usage is inefficient since the entire
Gram matrix is required to load into the main memory.
\item We propose new clustering assignment strategy which can reduce the
clustering assignment for $N$ samples in the entire training set
to the same task for $M$ equilibrium points where $M$ is usually
very small by a wide margin comparing to $N$.
\end{itemize}
The rest of this paper is organized as follows. In Section 2, we recall
large margin one-class support vector machine \cite{Nguyen_etal_14Kernel,Le2015}.
In Section 3, we introduce the budgeted stochastic gradient approach
to solve the optimization problem of large margin one-class support
vector machine in its primal form. In Section 4, we present our proposed
clustering assignment. Finally, in Section 5, we conduct the extensive
experiments and then discuss on the experimental results. In addition,
all proofs are given in the appendix section.

\section{Large margin one-class support vector machine}

Given the dataset $\mathcal{D}=\left\{ x_{1},x_{2},\ldots,x_{N}\right\} $,
to define the domain of novelty, we construct an optimal hyperplane
that can separate the data samples and the origin such that the margin,
i.e., the distance from the origin to the hyperplane, is maximized
\cite{Van2014,Le2015}. The optimization problem is formulated as
\begin{gather*}
\underset{\bw,\rho}{\max}\left(\frac{\left|\rho\right|}{\Vert\bw\Vert^{2}}\right)\\
\text{s.t.}:\,\transp{\bw}\phi(x_{i})-\rho\ge0,\,i=1,\ldots,N\\
\transp{\bw}\bzero-\rho=-\rho<0
\end{gather*}
where $\phi$ is a transformation from the input space to the feature
space and $\transp{\bw}\phi\left(x\right)-\rho=0$ is equation of
the hyperplane.

It occurs that the margin is invariant if we scale $(\bw,\rho)$ by
a factor $k$. Hence without loss of generality, we can assume that
$\rho=1$ and we achieve the following optimization problem
\begin{gather*}
\underset{\bw}{\text{min}}\left(\frac{1}{2}\Vert\bw\Vert^{2}\right)\\
\text{s.t.}:\,\transp{\bw}\phi(x_{i})-1\ge0,\,i=1,\ldots,N
\end{gather*}

Using the slack variables, we can extend the above optimization problem
to form the soft model of large margin one-class support vector machine
(LM-OCSVM)
\begin{gather*}
\underset{\bw}{\text{min}}\left(\frac{1}{2}\Vert\bw\Vert^{2}+\frac{C}{N}\sum_{i=1}^{N}\xi_{i}\right)\\
\text{s.t.}:\,\transp{\bw}\phi(x_{i})-1\ge-\xi_{i},\,i=1,\ldots,N\\
\xi_{i}\geq0,\,i=1,...,N
\end{gather*}
where $C>0$ is the trade-off parameter and $\bxi=\left[\xi_{1},\ldots,\xi_{N}\right]$
is the vector of slack variables.

We can rewrite the above optimization problem in the primal form as
follows
\begin{equation}
\underset{\bw}{\textrm{min}}\left(J(\bw)\triangleq\frac{1}{2}\left\Vert \bw\right\Vert ^{2}+\frac{C}{N}\sum_{i=1}^{N}\max\left\{ 0,1-\transp{\bw}\phi\left(x_{i}\right)\right\} \right)\label{eq:primal}
\end{equation}

\section{Budgeted stochastic gradient descent large margin one-class support
vector machine}

\subsection{Stochastic gradient descent solution}

We can apply the stochastic gradient descent (SGD) framework to solve
the optimization problem in Eq. (\ref{eq:primal}) in its primal form.
At the iteration $t$, we do the following steps:
\begin{itemize}
\item Uniformly sample $n_{t}$ from $[N]\triangleq\left\{ 1,2,\ldots,N\right\} $ 
\item Construct the instantaneous objective function 
\[
\mathcal{J}_{t}\left(\bw\right)=\frac{1}{2}\norm{\bw}^{2}+C\max\left\{ 0,1-\transp{\bw}\phi\left(x_{n_{t}}\right)\right\} 
\]
 
\item Set $g_{t}=\mathcal{J}_{t}^{'}\left(\bw_{t}\right)=\bw_{t}-C\mathbb{I}_{\transp{\bw_{t}}\phi\left(x_{n_{t}}\right)<1}\phi\left(x_{n_{t}}\right)$
where $\mathbb{I}_{A}$ is the indicator function, and is $1$ if
$A$ is true and is $0$ if otherwise
\item Update $\bw_{t+1}=\bw_{t}-\eta_{t}g_{t}=\frac{t-1}{t}\bw_{t}+C\eta_{t}\mathbb{I}_{\transp{\bw}\phi\left(x_{n_{t}}\right)<1}\phi\left(x_{n_{t}}\right)$
where $\eta_{t}=\frac{1}{t}$ is the learning rate
\end{itemize}
\begin{algorithm}
\caption{The pseudocode of stochastic gradient descent large margin one-class
support vector machine. \label{alg:SGD-LMOC}}

\textbf{Input}: $\mathcal{D}$, $K\left(.,.\right)$, $C$
\begin{algor}[1]
\item [{{*}}] $\bw_{1}=\bzero$
\item [{for}] $t=1$ \textbf{to} $T$

\begin{algor}[1]
\item [{{*}}] Sample $n_{t}\sim\text{Uniform}\left([N]\right)$
\item [{{*}}] $\bw_{t+1}=\frac{t-1}{t}\bw_{t}+C\eta_{t}\mathbb{I}_{\transp{\bw}\phi\left(x_{n_{t}}\right)<1}\phi\left(x_{n_{t}}\right)$
\end{algor}
\item [{endfor}]~
\end{algor}
\textbf{Output}: $\bw_{T+1}$
\end{algorithm}

The key steps of applying SGD to solve the optimization of LM-OCSVM
in its primal form (cf. Eq. (\ref{eq:primal})) is summarized in Algorithm
\ref{alg:SGD-LMOC}. At the iteration $t$, we uniformly sample a
data instance from the training set and then update the model using
this data instance. This kind of update enables the proposed algorithm
to be performed in online mode and encourages an economic memory usage.

\subsection{Budgeted stochastic gradient descent}

The aforementioned update rule is vulnerable to the curse of kernelization,
that is, the model size almost linearly grows up with the data size
accumulated over time. This makes the computation is gradually slower
across iterations and may cause potential memory overflow. To address
this issue, we use the budget approach wherein a budget $B$ is established
and when the current model size exceeds this budget, a budget maintenance
procedure is invoked to maintain the model size.

Algorithm \ref{alg:Algorithm-for-SGD-LMOCSVM} is proposed for BSGD-based
Large Margin One-class Support Vector Machine (BSGD-LMOC). It is noteworthy
that in Algorithm \ref{alg:Algorithm-for-SGD-LMOCSVM}, we store $\bw_{t}$
as $\sum_{i}\alpha_{i}\phi\left(x_{i}\right)$. Whenever the current
model size $b$ exceeds the budget $B$, we invoke the procedure $BM\left(\bw_{t+1},I_{t+1}\right)$
to perform the budget maintenance so as to maintain the current model
size \textbf{$b$ }to the budget $B$. Here we note that $I_{t+1}$
specifies the set of support indices during the iteration $t$ (cf.
Algorithm \ref{alg:Algorithm-for-SGD-LMOCSVM}).

\begin{algorithm}
\caption{Algorithm for BSGD-LMOC\label{alg:Algorithm-for-SGD-LMOCSVM}}

\textbf{Input}: $C,\,B,\,K\left(.,.\right)$
\begin{algor}[1]
\item [{{*}}] $\bw_{1}=\bzero$
\item [{{*}}] $b=0$
\item [{{*}}] $I_{1}=\emptyset$
\item [{for}] $t=1$ \textbf{to} $T$ 

\begin{algor}[1]
\item [{{*}}] Sample $n_{t}$ from $[N]$
\item [{{*}}] Update $\bw_{t+1}=\frac{t-1}{t}\bw_{t}+\alpha_{t}\phi\left(x_{n_{t}}\right)$
\hspace*{\fill}//$\alpha_{t}=C\eta_{t}\mathbb{I}_{y_{n_{t}}\transp{\bw_{t}}\phi\left(x_{n_{t}}\right)<\theta_{n_{t}}}y_{n_{t}}$
\item [{if}] $n_{t}\notin I_{t}$ and $\alpha_{t}\neq0$

\begin{algor}[1]
\item [{{*}}] $b=b+1$
\item [{{*}}] $I_{t+1}=I_{t}\,\bigcup\,$\{$n_{t}$\}
\item [{if}] $b>B$

\begin{algor}[1]
\item [{{*}}] BM$\left(\bw_{t+1},I_{t+1}\right)$\hspace*{\fill}//Budget
Maintenance
\item [{{*}}] $b=B$
\end{algor}
\item [{endif}]~
\end{algor}
\item [{endif}]~
\end{algor}
\item [{endfor}]~
\end{algor}
\textbf{Output}: $\bw_{T+1}$
\end{algorithm}

\subsection{Budget maintenance strategies \label{subsec:Budget-maintenance-strategies}}

In what follows, we present two budget maintenance strategies used
in our article (i.e., removal and projection). The first strategy
(i.e., removal) simply removes the most redundant vector. The second
one projects the most redundant vector onto the linear span of the
remaining vectors in the support set in the feature space before removing
it. We assume that
\[
\bw_{t+1}=\sum_{i\in I_{t+1}}\alpha_{i}\phi\left(x_{i}\right)
\]
The most redundant vector is determined as that with smallest coefficient
as 
\[
p_{t}=\argmin{i\in I_{t+1}}\left|\alpha_{i}\right|K\left(x_{i},x_{i}\right)
\]

\subsubsection{Removal}

The most redundant vector $\phi\left(x_{p_{t}}\right)$ is simply
removed. It follows that 
\[
\bw_{t+1}=\bw_{t}-\alpha_{p_{t}}\phi\left(x_{p_{t}}\right)
\]

\subsubsection{Projection}

The full projection version used in \cite{wang2012} requires to invert
a $B$ by $B$ matrix which costs $\text{O}\left(B^{3}\right)$. To
reduce the incurred computation, we propose to project the most redundant
vector $\phi\left(x_{p_{t}}\right)$ onto the set of $k$ vectors
in the feature space. This set of $k$ vectors can be determined by
either $k$ nearest neighbor of $x_{p_{t}}$ (i.e., $kNN\left(x_{p_{t}}\right)$)
in the input space or $k$ random vectors (i.e., $kRD\left(x_{p_{t}}\right)$)
in the support set. The model $\bw_{t+1}$ is incremented by the projection
of $\phi\left(x_{p_{t}}\right)$ onto the linear span of this set
in the feature space. Finally, the most redundant vector is removed.
In particular, we proceed as follows
\begin{gather*}
kInd=kNN\left(x_{p_{t}},I_{t+1}\right)\,\text{\,or}\,\,kRD\left(x_{p_{t}},I_{t+1}\right)\\
P_{t}=P\left(\phi\left(x_{p_{t}}\right),\text{span}\left(\phi\left(x_{i}\right):i\in kInd\right)\right)\\
\bw_{t+1}=\bw_{t}+\alpha_{p_{t}}P_{t}-\alpha_{p_{t}}\phi\left(x_{p_{t}}\right)
\end{gather*}
It is noteworthy that the computational cost to find the projection
$P_{t}$ is $\text{O}\left(k^{3}\right)$ where $k$ is usually a
small number (e.g., $k=5$) comparing with the cost $\text{O}\left(B^{3}\right)$
as in \cite{wang2012}. Since the projection has the form of $P_{t}=\sum_{i\in kInd}d_{i}\phi\left(x_{i}\right)$,
the model size is recovered to \textbf{$B$ }after performing the
budget maintenance strategy. 

\subsection{Convergence analysis}

In this section, we present the convergence analysis regarding the
convergence rate of $\bw_{t}$ to the optimal solution $\bw^{*}$.
Without loss of generality, we assume that data are bounded in the
feature space, i.e., $\norm{\phi\left(x\right)}\leq R,\,\forall x\in\mathcal{X}$.
We now introduce the theoretical results below and shall leave all
proofs to the appendix section. For comprehensibility, we make analysis
for the removal case. Let us define $Z_{t}$ to be the Bernoulli random
variable which specifies if the budget maintenance is performed at
round $t$. The update rule is as follows
\[
\bw_{t+1}=\bw_{t}-\eta_{t}g_{t}-Z_{t}\alpha_{p_{t}}\phi\left(x_{p_{t}}\right)
\]

Lemmas \ref{lem:st}, \ref{lem:wt}, \ref{lem:gt}, \ref{lem:remove},
and \ref{lem:wtw} establish the upper bound for the necessary quantities
that need for the analysis. The main theorem \ref{thm:regret} shows
the upper bound on the regret.
\begin{lem}
\label{lem:st}Let us denote $s_{t}=\sum_{i\in I_{t}}\left|\alpha_{i}\right|$.
We then have for all $t$,
\[
s_{t}\leq CR
\]
\end{lem}

\begin{lem}
\label{lem:wt}The following statement holds for all $t$,
\[
\norm{\bw_{t}}\le CR^{2}
\]
\end{lem}

\begin{lem}
\label{lem:gt}The following statement holds for all $t$,
\[
\norm{g_{t}}\leq G=CR+CR^{2}
\]
\end{lem}

\begin{lem}
\label{lem:remove}Given a positive number $m$, assume that before
removing $\left(x_{p_{t}},y_{p_{t}}\right)$ is updated at most $m$
times. We then have the following statement for all $t$,
\[
\left|\alpha_{p_{t}}\right|\leq\frac{mCR}{t}
\]
\end{lem}

\begin{lem}
\label{lem:ht}Let us denote $\rho_{i}=\frac{\alpha_{i}}{\eta_{t}}=t\alpha_{i},\,\forall i\in I_{t}$
and $h_{t}=Z_{t}\rho_{p_{t}}\phi\left(x_{p_{t}}\right)$. We then
have
\[
\norm{h_{t}}\leq H=mCR^{2}
\]
\end{lem}

\begin{lem}
\label{lem:wtw}We have the following inequality for all $t$,
\[
\mathbb{E}\left[\norm{\bw_{t}-\bw^{*}}^{2}\right]\leq W^{2}=\left(H+\sqrt{H^{2}+\left(G+H\right)^{2}}\right)^{2}
\]
\end{lem}

\begin{thm}
\label{thm:regret} Considering the running of Algorithm \ref{alg:Algorithm-for-SGD-LMOCSVM},
we have the following inequality
\begin{gather*}
\mathbb{E}\left[\mathcal{J}\left(\overline{\bw}_{T}\right)\right]-\mathcal{J}\left(\bw^{*}\right)\leq\frac{1}{T}\sum_{t=1}^{T}\mathbb{E}\left[\mathcal{J}\left(\bw_{t}\right)\right]-\mathcal{J}\left(\bw^{*}\right)\\
\leq\frac{\left(G+H\right)^{2}\left(\log T+1\right)}{2T}+\frac{WR}{T}\sum_{t=1}^{T}\mathbb{P}\left(Z_{t}=1\right)\mathbb{E}\left[\rho_{p_{t}}^{2}\right]^{1/2}
\end{gather*}
\end{thm}

Theorem \ref{thm:regret} reveals that there exists an error gap between
the outputting solution and the optimal solution. This gap crucially
depends on the budget maintenance rate $\mathbb{P}\left(Z_{t}=1\right)$
and the gradient error $\rho_{p_{t}}=\frac{\alpha_{p_{t}}}{\eta_{t}}$.
It also discloses that we need to remove the vector with smallest
absolute coefficient (i.e., $\left|\alpha_{p_{t}}\right|$) to rapidly
reduce the error gap.

It is noteworthy that our analysis also covers the standard analysis.
In particular, when the budget maintenance never happens (i.e., $\mathbb{P}\left(Z_{t}=1\right)=0$),
we gain the typical convergence rate $\text{O}\left(\frac{\log\,T}{T}\right)$.

\section{Clustering assignment}

After solving the optimization problem, we yield the decision function
\[
f\left(x\right)=\sum_{i=1}^{N}\alpha_{i}K\left(x_{i},x\right)-1
\]
To find the equilibrium points, we need to solve the equation $\nabla f\left(x\right)=0$.
To this end, we use the fixed point technique and assume that Gaussian
kernel is used, i.e., $K\left(x,x'\right)=e^{-\gamma\Vert x-x'\Vert^{2}}$.
We then have
\[
\frac{1}{2}\nabla f\left(x\right)=\sum_{i=1}^{N}\alpha_{i}\left(x_{i}-x\right)e^{-\gamma\Vert x-x_{i}\Vert^{2}}=0\goto x=\frac{\sum_{i=1}^{N}\alpha_{i}e^{-\gamma\Vert x-x_{i}\Vert^{2}}x_{i}}{\sum_{i=1}^{N}\alpha_{i}e^{-\gamma\Vert x-x_{i}\Vert^{2}}}=P\left(x\right)
\]

To find an equilibrium point, we start with the initial point $x^{(0)}\in\mathbb{R}^{d}$
and iterate $x^{(j+1)}=P\left(x^{(j)}\right)$. By fixed point theorem,
the sequence $x^{(j)}$, which can be considered as a trajectory with
start $x^{(0)}$, converges to the point $x_{*}^{(0)}$ satisfying
P$\left(x_{*}^{(0)}\right)=x_{*}^{(0)}$ or $\nabla f\left(x_{*}^{(0)}\right)=0$,
i.e., $x_{*}^{(0)}$ is an equilibrium point.

Let us denote $B_{\epsilon}=\left\{ x_{i}:1\leq i\leq N\,\wedge\,\vert f\left(x_{i}\right)\vert\leq\epsilon\right\} $,
namely the extended boundary for a tolerance $\epsilon>0$. It follows
that the set $B_{\epsilon}$ forms a strip enclosing the decision
boundary $f\left(x\right)=0$. Algorithm \ref{alg:Clustering-assignment}
is proposed to do clustering assignment. In Algorithm \ref{alg:Clustering-assignment},
the task of clustering assignment is reduced to itself for $M$ equilibrium
points. To fulfill cluster assignment for $M$ equilibrium points,
we run $m=20$ sample-point test as proposed in \cite{Ben-Hur01supportvector}.

\begin{algorithm}[h]
\caption{Clustering assignment procedure.\label{alg:Clustering-assignment}}

\textbf{Input}: $f\left(x\right)=\sum_{i=1}^{N}\alpha_{i}K\left(x_{i},x\right)-1$,
$B_{\varepsilon}$
\begin{algor}[1]
\item [{{*}}] $E=\emptyset$
\item [{for}] $x^{(0)}$ \textbf{in} $B_{\epsilon}$

\begin{algor}[1]
\item [{{*}}] Find the equilibrium point $x_{*}^{(0)}$
\end{algor}
\item [{if}] $x_{*}^{(0)}\notin E$

\begin{algor}[1]
\item [{{*}}] $E=E\,\cup$\{$x_{*}^{(0)}$\}
\end{algor}
\item [{endif}]~
\item [{endfor}]~
\item [{{*}}] Do $m$ sample point test for $E$ to find cluster indices
for $e_{1},e_{2},\ldots,e_{M}$\\
//\textit{Assume that $E=\left\{ e_{1},e_{2},\ldots,e_{M}\right\} $}
\item [{{*}}] Each point $x^{(0)}\in B_{\epsilon}$ is assigned to the
cluster of its corresponding equilibrium point $x_{*}^{(0)}\in E$
\item [{{*}}] Each point $x\in\mathcal{D}\backslash B_{\epsilon}$ is assigned
to the cluster of its nearest neighbor in $B_{\epsilon}$ using the
Euclidean distance
\end{algor}
\textbf{Output}: clustering solution for $\mathcal{D=}$\{$x_{1},\ldots,x_{N}$\}
\end{algorithm}

Our proposed clustering assignment procedure is different from the
existing procedure proposed in \cite{Ben-Hur01supportvector}. The
procedure proposed in \cite{Ben-Hur01supportvector} requires to run
$m=20$ sample-point test for every edge connected $x_{i},\,x_{j}$
($i\neq j$) in the training set. Consequently, the computational
cost incurred is $\text{O}\left(N\left(N-1\right)ms\right)$ where
$s$ is the sparsity level of the decision function (i.e., the number
of vectors in the model). Our proposed procedure needs to perform
$m=20$ sample-point test for a reduced set of $M$ data samples (i.e.,
the set of the equilibrium points $\left\{ e_{1},e_{2},\ldots,e_{M}\right\} $)
where $M$ is possibly very small comparing with $N$. The reason
is that many data points in the training set could converge to a common
equilibrium point which significantly reduces the size from $N$ to
$M$. The computational cost incurred is therefore $\text{O}\left(M\left(M-1\right)ms\right)$.

\section{Experiments}

\subsection{Visual experiment}

\begin{figure}[h]
\begin{centering}
\begin{tabular}{ccc}
\includegraphics[width=0.3\textwidth]{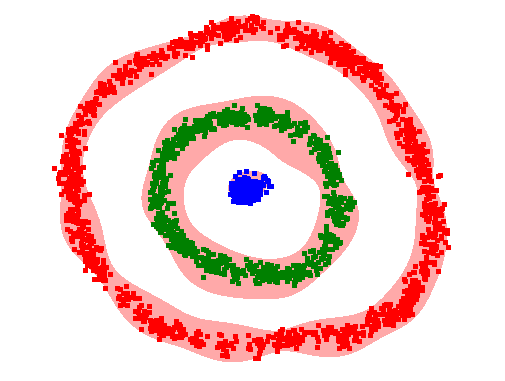} & \includegraphics[width=0.3\textwidth]{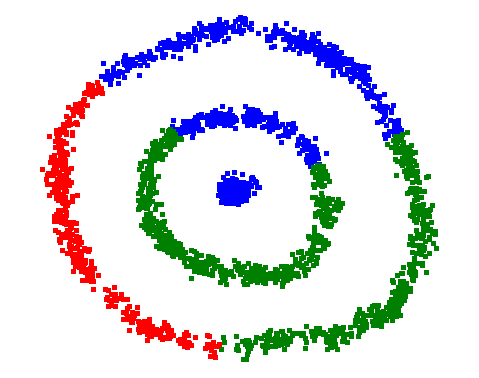} & \includegraphics[width=0.3\textwidth]{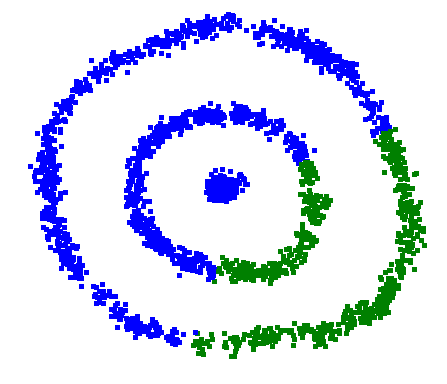}\tabularnewline
The proposed method & C-Means & Fuzzy C-Means\tabularnewline
\end{tabular}
\par\end{centering}
\caption{Visual comparison of the proposed method (the orange region is the
domain of novelty) with C-Means and Fuzzy C-Means on two ring dataset.
\label{fig:nested}}
\end{figure}

\begin{figure}[h]
\begin{centering}
\begin{tabular}{ccc}
\includegraphics[width=0.3\textwidth]{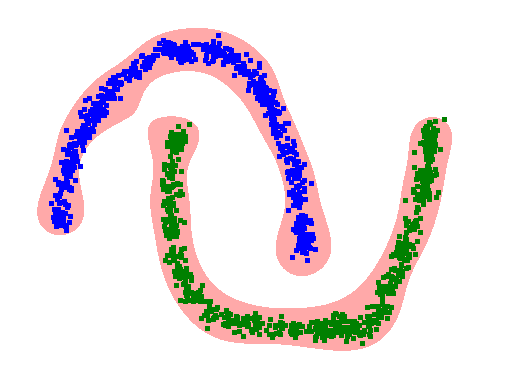} & \includegraphics[width=0.3\textwidth]{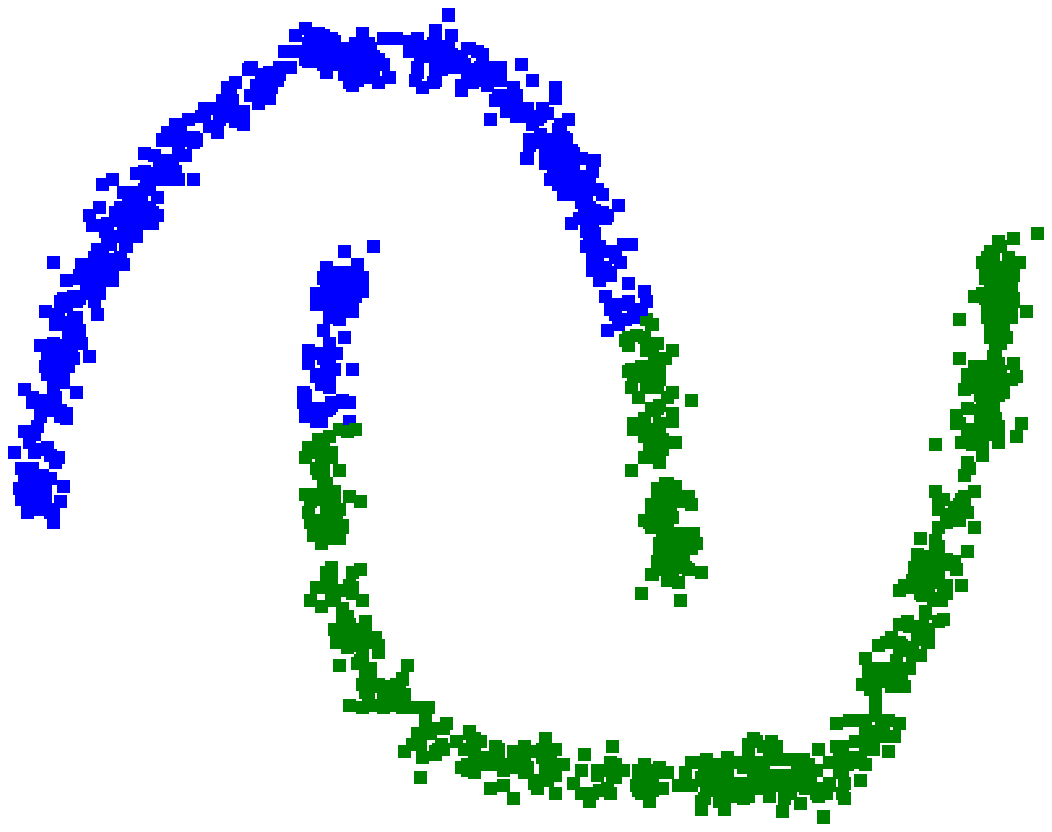} & \includegraphics[width=0.3\textwidth]{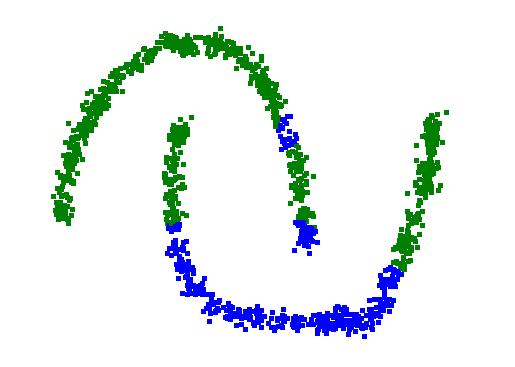}\tabularnewline
The proposed method & C-Means & Fuzzy C-Means\tabularnewline
\end{tabular}
\par\end{centering}
\caption{Visual comparison of the proposed method (the orange region is the
domain of novelty) with C-Means and Fuzzy C-Means on two moon dataset.
\label{fig:2moon}}
\end{figure}

\begin{figure}[h]
\begin{centering}
\includegraphics[width=0.4\textwidth]{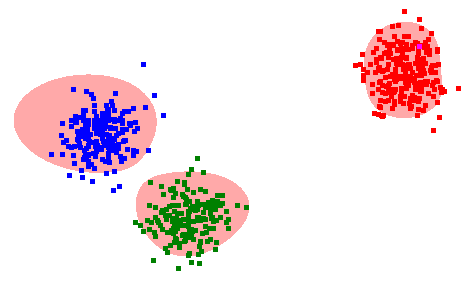}~~~~~~~~~~~~~\includegraphics[width=0.4\textwidth]{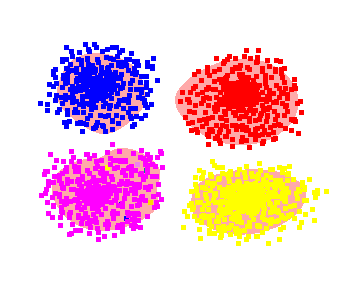}
\par\end{centering}
\centering{}\caption{The proposed method (the orange region is the domain of novelty) can
recognize the clusters scattered from mixture of either $3$ (left)
or $4$ (right) Gaussian distributions. \label{fig:4gauss}}
\end{figure}

To visually show the high clustering quality produced by our proposed
method using budgeted SGD in conjunction with the removal strategy,
we establish experiment on three synthesized datasets and visually
make comparison of the proposed method with C-Means and Fuzzy C-Means.
In the first experiment, data samples form the nested structure with
two outside rings and one Gaussian distribution at center. As shown
in Fig. \ref{fig:nested}, our method can perfectly detect three clusters
without any prior information whilst both C-Means and Fuzzy C-Means
with the number of clusters being set to $3$ beforehand fail to discover
the nested clusters. The second experiment is carried out with a two-moon
dataset. As observed from Fig. \ref{fig:2moon}, our method without
any prior knowledge can perfectly discover two clusters in moons.
However, C-Means and Fuzzy C-Means cannot detect the clusters correctly.
In the last visual experiment, we generate data from the mixture of
$3$ or $4$ Gaussian distributions. As shown in Fig. \ref{fig:4gauss},
our method can perfectly detect $3$ (left) and $4$ (right) clusters
corresponding to the individual Gaussian distributions. These visual
experiments demonstrate that the proposed method is able to not only
generate the cluster boundaries in arbitrary shapes but also automatically
detect the appropriate number of clusters well presented the data.

\subsection{Experiment on real datasets}

To explicitly prove the performance of the proposed algorithm, we
conduct experiments on the real datasets. Cluster analysis is naturally
an unsupervised learning task and therefore, there does not exist
a perfect measure to compare given two clustering algorithms. We make
use of five typical clustering validity indices (CVI) including compactness,
purity, rand index, Davies-Bouldin index (DB index), and normalized
mutual information (NMI) for comparison. A good clustering algorithm
should achieve a solution which has a high purity, rand index, DB
index, and NMI and a low compactness. 

\subsubsection*{Clustering validity index}

Compactness measures the average pairwise distances of points in the
same cluster \cite{Halkidi:2002} and is given as follows

\[
Compactness\triangleq\frac{1}{N}\sum_{k=1}^{m}N_{k}\frac{\sum_{x,x'\in C_{k}}d\left(x,x'\right)}{N_{k}\left(N_{k}-1\right)/2}
\]

\noindent where the cluster solution consists of $m$ clusters $C_{1},\,C_{2},\,...,C_{m}$
whose cardinalities are $N_{1},\,N_{2},\,...,\,N_{m}$, respectively.

The clustering with a small compactness is preferred. A small compactness
gained means the average intra-distance of clusters is small and homogeneity
is thereby good, i.e., two objects in the same cluster has high similarity
to each other. 

The second CVI in use is purity which measures the purity of clustering
solution with respect to the nature classes of data \cite{Murphy:2012}.
It is certainly true that the metric purity is only appropriate for
data with labels in nature. Let $N_{ij}$ be the number of objects
in cluster $i$ that belong to the class $j$. Again, let $N_{i}\triangleq\sum_{j=1}^{m}N_{ij}$
be total number of objects in cluster $i$. Let us define $p_{ij}\triangleq\frac{N_{ij}}{N_{j}}$,
i.e., the empirical distribution over class labels for cluster $i$.
We define a purity of a cluster as $p_{i}\triangleq\underset{j}{\text{max}}\,p_{ij}$
and overall purity of a clustering solution as
\[
Purity\triangleq\sum_{i}\frac{N_{i}}{N}\times p_{i}
\]

The purity ranges between 0 (bad) and 1 (good). This CVI embodies
the classification ability of clustering algorithm. A clustering algorithm
which achieves a high purity can be appropriately used for classification
purpose.

The third CVI used as a measure is rand index \cite{Murphy:2012}.
To calculate this CVI for a clustering solution, we need to construct
a $2\times2$ contingency table containing the following numbers:
1) TP (true positive) is the number of pairs that are in the same
cluster and belong to the same class; 2) TN (true negative) is the
number of pairs that are in two different clusters and belong to different
classes; 3) FP (false positive) is the number of pairs that are in
the same cluster but belong to different classes; and 4) FN (false
negative) is the number of pairs that are in two different clusters
but belong to the same class. Rand index is defined as follows

\[
Rand\triangleq\frac{TP+TN}{TP+FP+TN+FN}
\]

This can be interpreted as the fraction of clustering decisions that
are correct. Obviously, rand index ranges between 0 and 1.

Davies-Bouldin validity index is a function of the ratio of the sum
of intra-distances to inter-distances \cite{Halkidi:2002} and is
formulated as follows

\[
DBI\triangleq\frac{1}{m}\sum_{i=1}^{m}\underset{j\neq i}{\text{max}}\left\{ \frac{\Delta\left(C_{i}\right)+\Delta\left(C_{j}\right)}{d\left(C_{i},C_{j}\right)}\right\} 
\]

A good clustering algorithm should produce the solution which has
as small DBI as possible.

The last considered CVI is normalized mutual information (NMI) \cite{Murphy:2012}.
This measure allows us to trade off the quality of the clustering
against the number of clusters. 
\[
NMI\triangleq\frac{I\left(\Omega,C\right)}{\left[H\left(C\right)+H\left(\Omega\right)\right]/2}
\]
where $C=\left\{ c_{1},...,c_{J}\right\} $ is the set of classes
and $\Omega=\left\{ \omega_{1},...,\omega_{K}\right\} $ is the set
of clusters.

$I\left(\Omega,C\right)$ is the mutual information and is defined
as 
\[
I\left(\Omega,C\right)\triangleq\sum_{k}\sum_{j}P\left(c_{j}\bigcap\omega_{k}\right)\log\frac{P\left(c_{j}\bigcap\omega_{k}\right)}{P\left(c_{j}\right)P\left(\omega_{k}\right)}
\]
and $H\left(.\right)$ is the entropy and is defined as 
\[
H\left(\Omega\right)\triangleq-\sum_{k}P\left(\omega_{k}\right)\log P\left(\omega_{k}\right)
\]

It is certainly that the NMI ranges between 0 and 1, and a good clustering
algorithm should produce as high NMI measure as possible.

We perform experiments on $15$ well-known datasets for clustering
task. The statistics of the experimental datasets is given in Table
\ref{tab:Details}. These datasets are fully labeled and consequently,
the CVIs like purity, rand index, and NMI can be completely estimated.
We make comparison of our proposed method with the following baselines.
\begin{table}[H]
\begin{centering}
\begin{tabular}{|c|c|c|c|}
\hline 
\textbf{Datasets} & \textbf{Size} & \textbf{Dimension} & \textbf{\#Classes}\tabularnewline
\hline 
Aggregation & 788 & 2 & 7\tabularnewline
\hline 
Breast Cancer & 699 & 9 & 2\tabularnewline
\hline 
Compound & 399 & 2 & 6\tabularnewline
\hline 
D31 & 3,100 & 2 & 31\tabularnewline
\hline 
Flame & 240 & 2 & 2\tabularnewline
\hline 
Glass & 214 & 9 & 7\tabularnewline
\hline 
Iris & 150 & 4 & 3\tabularnewline
\hline 
Jain & 373 & 2 & 2\tabularnewline
\hline 
Pathbased & 300 & 2 & 3\tabularnewline
\hline 
R15 & 600 & 2 & 15\tabularnewline
\hline 
Spiral & 312 & 2 & 3\tabularnewline
\hline 
Abalone & 4,177 & 8 & 28\tabularnewline
\hline 
Car & 1,728 & 6 & 4\tabularnewline
\hline 
Musk & 6,598 & 198 & 2\tabularnewline
\hline 
Shuttle & 43,500 & 9 & 5\tabularnewline
\hline 
\end{tabular}
\par\end{centering}
\caption{The statistics of the experimental datasets. \label{tab:Details}}
\end{table}

\subsubsection*{Baselines}
\begin{itemize}
\item \textit{Support vector clustering (SVC)}\textbf{\textit{ }}\cite{Ben-Hur01supportvector}:
using LIBSVM \cite{libsvm} to find domain of novelty and fully connected
graph for clustering assignment.
\item \textit{Fast support vector clustering (FSVC)} \cite{JungLL10}: an
equilibrium-based approach for clustering assignment.
\item \emph{SGD Large Margin One-class Support Vector Machine (SGD-LMOC)}
\cite{pham2016fast}: the SGD-based solution of Large Margin One-class
Support Vector Machine which is vulnerable to the curse of kernelization. 
\end{itemize}
Our proposed method has three variants which correspond to three budget
maintenance strategies. We name these three variants as BSGD-R (BSGD
+ removal strategy), BSGD-PNN (BSGD + projection strategy with $k$-nearest
neighbors), and BSGD-PRN (BSGD + projection strategy with $k$-random
neighbors). It is noteworthy that we use BSGD-R, BSGD-PNN, or BSGD-PRN
(cf. Algorithm \ref{alg:Algorithm-for-SGD-LMOCSVM} and Section \ref{subsec:Budget-maintenance-strategies})
to train the model in the first phase and then use Algorithm \ref{alg:Clustering-assignment}
to perform clustering assignment in the second phase. All competitive
methods are run on a Windows computer with dual-core CPU $2.6\text{GHz}$
and $4\text{GB}$ RAM.

\subsubsection*{Hyperparameter Setting}

The RBF kernel, given by $K\left(x,x'\right)=e^{-\gamma\left\Vert x-x'\right\Vert ^{2}}$,
is employed. The width of kernel $\gamma$ is searched on the grid
$\left\{ 2^{-5},\,2^{-3},\,\ldots,\,2^{3},\,2^{5}\right\} $. The
trade-off parameter $C$ is searched on the same grid. In addition,
the parameters $p$ and $\varepsilon$ in FSVC are searched in the
common grid $\left\{ 0.1,0.2,\ldots,0.9,1\right\} $ which is the
same as in \cite{JungLL10}. Determining the number of iterations
in Algorithm \ref{alg:Algorithm-for-SGD-LMOCSVM} is really challenging.
To resolve it, we use the stopping criterion $\left\Vert \bw_{t+1}-\bw_{t}\right\Vert \leq\theta=0.01$
(i.e., the next hyperplane does only a slight change). For the projection
strategy in BSGD-PNN and BSGD-PRN, we set $k=5$ to efficiently relax
the matrix inversion.

\subsubsection*{Experimental Result}

\begin{table}[H]
\begin{centering}
\resizebox{0.95\textwidth}{!}{%
\begin{tabular}{|c|c|c|c|c|c|c|}
\hline 
\multirow{2}{*}{\textbf{\small{}Dataset}} & \multicolumn{6}{c|}{\textbf{Purity}}\tabularnewline
\cline{2-7} 
 & \textbf{\small{}SVC} & \textbf{\small{}SGD} & \textbf{FSVC} & \textbf{BSGD-R} & \textbf{BSGD-PNN} & \textbf{BSGD-PRN}\tabularnewline
\hline 
Aggregation {[}50{]}  & \textbf{1.00} & \textbf{1.00} & 0.22 & \textbf{1.00} & \textbf{1.00} & \textbf{1.00}\tabularnewline
\hline 
Breast Cancer {[}50{]} & 0.98 & \textbf{0.99} & \textbf{0.99} & 0.95 & \textbf{0.99} & 0.97\tabularnewline
\hline 
Compound {[}50{]} & 0.66 & 0.62 & 0.13 & \textbf{0.99} & 0.98 & 0.97\tabularnewline
\hline 
Flame {[}50{]} & 0.86 & 0.87 & 0.03 & \textbf{1.00} & \textbf{1.00} & \textbf{1.00}\tabularnewline
\hline 
Glass {[}50{]} & 0.50 & 0.71 & 0.65 & 0.88 & 0.87 & \textbf{0.93}\tabularnewline
\hline 
Iris {[}50{]} & \textbf{1.00} & \textbf{1.00} & 0.68 & \textbf{1.00} & \textbf{1.00} & 0.99\tabularnewline
\hline 
Jain {[}50{]} & 0.37 & 0.46 & 0.69 & \textbf{1.00} & \textbf{1.00} & \textbf{1.00}\tabularnewline
\hline 
Pathbased {[}50{]} & 0.60 & 0.50 & \textbf{1.00} & \textbf{1.00} & \textbf{1.00} & \textbf{1.00}\tabularnewline
\hline 
R15 {[}50{]} & 0.88 & 0.90 & 0.37 & \textbf{1.00} & 0.99 & \textbf{1.00}\tabularnewline
\hline 
Spiral {[}50{]} & 0.09 & 0.33 & 0.53 & \textbf{1.00} & \textbf{1.00} & \textbf{1.00}\tabularnewline
\hline 
D31 {[}50{]} & 0.94 & \textbf{0.99} & 0.42 & 0.96 & 0.97 & 0.91\tabularnewline
\hline 
Abalone {[}50{]} & 0.22 & 0.44 & 0.03 & 0.33 & \textbf{0.57} & 0.36\tabularnewline
\hline 
Car {[}50{]} & 0.94 & \textbf{0.95} & 0.70 & 0.91 & 0.83 & 0.86\tabularnewline
\hline 
Musk {[}50{]} & 0.87 & 0.68 & \textbf{0.88} & 0.42 & 0.64 & 0.60\tabularnewline
\hline 
Shuttle {[}100{]} & 0.06 & 0.05 & 0.06 & 0.34 & \textbf{0.90} & 0.42\tabularnewline
\hline 
\end{tabular}}
\par\end{centering}
\caption{The purity of the clustering methods on the experimental datasets.
Larger is better. The number (i.e., {[}x{]}) nearby the dataset name
specifies the budget size used in the proposed method. \label{tab:pu_rand_nmi}}
\end{table}

\begin{table}[H]
\begin{centering}
\resizebox{0.95\textwidth}{!}{%
\begin{tabular}{|c|c|c|c|c|c|c|}
\hline 
\multirow{2}{*}{\textbf{\small{}Dataset}} & \multicolumn{6}{c|}{\textbf{Rand Index}}\tabularnewline
\cline{2-7} 
 & \textbf{\small{}SVC} & \textbf{\small{}SGD} & \textbf{FSVC} & \textbf{BSGD-R} & \textbf{BSGD-PNN} & \textbf{BSGD-PRN}\tabularnewline
\hline 
Aggregation {[}50{]}  & \textbf{1.00} & \textbf{1.00} & 0.22 & 0.94 & 0.93 & 0.95\tabularnewline
\hline 
Breast Cancer {[}50{]} & 0.82 & \textbf{0.85} & 0.81 & 0.73 & 0.78 & 0.72\tabularnewline
\hline 
Compound {[}50{]} & 0.92 & 0.88 & 0.25 & 0.90 & \textbf{0.95} & 0.91\tabularnewline
\hline 
Flame {[}50{]} & 0.75 & 0.76 & 0.03 & 0.87 & 0.74 & \textbf{0.91}\tabularnewline
\hline 
Glass {[}50{]} & 0.77 & \textbf{0.91} & 0.54 & 0.78 & 0.83 & 0.81\tabularnewline
\hline 
Iris {[}50{]} & \textbf{0.97} & 0.96 & 0.69 & 0.83 & 0.84 & 0.81\tabularnewline
\hline 
Jain {[}50{]} & 0.70 & 0.71 & 0.77 & \textbf{1.00} & 0.98 & 0.89\tabularnewline
\hline 
Pathbased {[}50{]} & 0.81 & \textbf{0.94} & 1.00 & 0.71 & 0.77 & 0.76\tabularnewline
\hline 
R15 {[}50{]} & 0.74 & 0.71 & 0.37 & 0.95 & 0.95 & \textbf{0.96}\tabularnewline
\hline 
Spiral {[}50{]} & 0.15 & \textbf{0.94} & 0.75 & 0.91 & 0.75 & 0.75\tabularnewline
\hline 
D31 {[}50{]} & 0.88 & 0.81 & 0.54 & \textbf{0.98} & \textbf{0.98} & 0.96\tabularnewline
\hline 
Abalone {[}50{]} & 0.43 & 0.86 & 0.12 & 0.88 & \textbf{0.89} & 0.88\tabularnewline
\hline 
Car {[}50{]} & 0.46 & 0.46 & 0.54 & 0.55 & 0.56 & \textbf{0.59}\tabularnewline
\hline 
Musk {[}50{]} & 0.26 & 0.28 & 0.26 & 0.74 & \textbf{0.76} & 0.75\tabularnewline
\hline 
Shuttle {[}100{]} & \textbf{0.84} & 0.83 & 0.75 & 0.50 & 0.62 & 0.63\tabularnewline
\hline 
\end{tabular}}
\par\end{centering}
\caption{The rand index of the clustering methods on the experimental datasets.
Larger is better. The number (i.e., {[}x{]}) nearby the dataset name
specifies the budget size used in the proposed method.\label{tab:rand}}
\end{table}

\begin{table}[H]
\begin{centering}
\resizebox{0.95\textwidth}{!}{%
\begin{tabular}{|c|c|c|c|c|c|c|}
\hline 
\multirow{2}{*}{\textbf{\small{}Dataset}} & \multicolumn{6}{c|}{\textbf{NMI}}\tabularnewline
\cline{2-7} 
 & \textbf{\small{}SVC} & \textbf{\small{}SGD} & \textbf{FSVC} & \textbf{BSGD-R} & \textbf{BSGD-PNN} & \textbf{BSGD-PRN}\tabularnewline
\hline 
Aggregation {[}50{]}  & 0.69 & 0.75 & 0.60 & \textbf{0.89} & \textbf{0.89} & \textbf{0.89}\tabularnewline
\hline 
Breast Cancer {[}50{]} & 0.22 & \textbf{0.55} & 0.45 & 0.42 & 0.43 & 0.41\tabularnewline
\hline 
Compound {[}50{]} & 0.51 & 0.81 & 0.45 & \textbf{0.82} & \textbf{0.82} & 0.77\tabularnewline
\hline 
Flame {[}50{]} & 0.55 & 0.51 & 0.05 & 0.57 & 0.54 & \textbf{0.66}\tabularnewline
\hline 
Glass {[}50{]} & \textbf{0.60} & 0.44 & 0.53 & 0.55 & 0.54 & 0.56\tabularnewline
\hline 
Iris {[}50{]} & 0.63 & 0.75 & 0.71 & \textbf{0.76} & \textbf{0.76} & \textbf{0.76}\tabularnewline
\hline 
Jain {[}50{]} & 0.53 & 0.31 & \textbf{1.00} & 0.98 & 0.92 & 0.68\tabularnewline
\hline 
Pathbased {[}50{]} & 0.48 & 0.43 & 0.12 & 0.49 & \textbf{0.58} & 0.52\tabularnewline
\hline 
R15 {[}50{]} & 0.67 & 0.77 & 0.77 & 0.80 & \textbf{0.83} & 0.80\tabularnewline
\hline 
Spiral {[}50{]} & 0.52 & 0.34 & 0.16 & \textbf{0.85} & 0.58 & 0.59\tabularnewline
\hline 
D31 {[}50{]} & 0.45 & 0.50 & 0.38 & 0.80 & 0.81 & \textbf{0.89}\tabularnewline
\hline 
Abalone {[}50{]} & 0.22 & \textbf{0.34} & 0.07 & 0.27 & 0.31 & 0.28\tabularnewline
\hline 
Car {[}50{]} & 0.32 & 0.32 & 0.24 & \textbf{0.33} & 0.33 & 0.29\tabularnewline
\hline 
Musk {[}50{]} & 0.21 & 0.16 & \textbf{0.23} & 0.08 & 0.14 & 0.11\tabularnewline
\hline 
Shuttle {[}100{]} & 0.26 & 0.41 & 0.50 & 0.38 & 0.47 & \textbf{0.52}\tabularnewline
\hline 
\end{tabular}}
\par\end{centering}
\caption{The NMI of the clustering methods on the experimental datasets. Larger
is better. The number (i.e., {[}x{]}) nearby the dataset name specifies
the budget size used in the proposed method.\label{tab:nmi}}
\end{table}

\begin{table}[H]
\begin{centering}
\resizebox{0.95\textwidth}{!}{%
\begin{tabular}{|c|c|c|c|c|c|c|}
\hline 
\multirow{2}{*}{\textbf{\small{}Dataset}} & \multicolumn{6}{c|}{\textbf{\small{}Compactness }}\tabularnewline
\cline{2-7} 
 & \textbf{\small{}SVC} & \textbf{\small{}SGD} & \textbf{FSVC} & \textbf{BSGD-R} & \textbf{BSGD-PNN} & \textbf{BSGD-PRN}\tabularnewline
\hline 
Aggregation {[}50{]}  & 0.29 & 0.29 & 2.84 & \textbf{0.01} & \textbf{0.01} & 0.02\tabularnewline
\hline 
Breast Cancer {[}50{]} & 1.26 & 0.68 & \textbf{0.71} & 2.17 & 1.04 & 2.18\tabularnewline
\hline 
Compound {[}50{]} & 0.5 & 0.21 & 2.43 & 0.02 & \textbf{0.01} & 0.03\tabularnewline
\hline 
Flame {[}50{]} & 0.58 & 0.44 & 2.28 & \textbf{0.03} & \textbf{0.03} & \textbf{0.03}\tabularnewline
\hline 
Glass {[}50{]} & 0.72 & 0.68 & 1.85 & 0.75 & 0.69 & \textbf{0.34}\tabularnewline
\hline 
Iris {[}50{]} & 0.98 & 0.25 & 0.99 & 0.02 & \textbf{0.01} & 0.05\tabularnewline
\hline 
Jain {[}50{]} & 0.96 & 0.36 & 1.16 & \textbf{0.01} & \textbf{0.01} & \textbf{0.01}\tabularnewline
\hline 
Pathbased {[}50{]} & 0.18 & 0.3 & 1.04 & \textbf{0.01} & \textbf{0.01} & \textbf{0.01}\tabularnewline
\hline 
R15 {[}50{]} & 0.61 & 0.13 & 1.84 & \textbf{0.01} & \textbf{0.01} & \textbf{0.01}\tabularnewline
\hline 
Spiral {[}50{]} & 2.00 & 0.17 & 0.18 & \textbf{0.01} & \textbf{0.01} & 0.02\tabularnewline
\hline 
D31 {[}50{]} & 1.41 & 0.26 & 1.78 & \textbf{0.01} & \textbf{0.01} & 0.25\tabularnewline
\hline 
Abalone {[}50{]} & 3.88 & 0.40 & 4.97 & 0.23 & \textbf{0.15} & 0.21\tabularnewline
\hline 
Car {[}50{]} & 0.75 & \textbf{0.74} & 14.68 & 1.32 & 3.06 & 5.25\tabularnewline
\hline 
Musk {[}50{]} & \textbf{9.89} & 30.05 & 20.00 & 100.16 & 31.17 & 40.42\tabularnewline
\hline 
Shuttle {[}100{]} & 0.50 & 0.46 & \textbf{0.26} & 0.28 & 0.50 & 0.50\tabularnewline
\hline 
\end{tabular}}
\par\end{centering}
\caption{The compactness of the clustering methods on the experimental datasets.
Smaller is better. The number (i.e., {[}x{]}) nearby the dataset name
specifies the budget size used in the proposed method.\label{tab:com}}
\end{table}

\begin{table}[H]
\begin{centering}
\resizebox{0.95\textwidth}{!}{%
\begin{tabular}{|c|c|c|c|c|c|c|}
\hline 
\multirow{2}{*}{\textbf{\small{}Dataset}} & \multicolumn{6}{c|}{\textbf{\small{}DB Index}}\tabularnewline
\cline{2-7} 
 & \textbf{\small{}SVC} & \textbf{\small{}SGD} & \textbf{FSVC} & \textbf{BSGD-R} & \textbf{BSGD-PNN} & \textbf{BSGD-PRN}\tabularnewline
\hline 
Aggregation {[}50{]}  & 0.68 & 0.67 & 0.63 & 4.45 & \textbf{4.53} & 3.99\tabularnewline
\hline 
Breast Cancer {[}50{]} & 1.58 & 1.38 & 0.53 & 4.80 & 4.64 & \textbf{5.71}\tabularnewline
\hline 
Compound {[}50{]} & 2.45 & 0.86 & 0.67 & 3.84 & \textbf{5.15} & 3.01\tabularnewline
\hline 
Flame {[}50{]} & 1.30 & 0.65 & 3.56 & 3.42 & \textbf{4.01} & 3.29\tabularnewline
\hline 
Glass {[}50{]} & 0.53 & 0.56 & 0.93 & 3.98 & \textbf{4.64} & 4.14\tabularnewline
\hline 
Iris {[}50{]} & 1.95 & 1.17 & 0.77 & 3.42 & \textbf{3.96} & 3.44\tabularnewline
\hline 
Jain {[}50{]} & 1.23 & 1.08 & 0.71 & 4.38 & 4.19 & \textbf{4.22}\tabularnewline
\hline 
Pathbased {[}50{]} & 0.36 & 0.73 & 1.07 & 4.02 & \textbf{4.42} & 3.71\tabularnewline
\hline 
R15 {[}50{]} & 2.96 & 1.42 & 1.37 & 4.00 & \textbf{4.33} & 3.91\tabularnewline
\hline 
Spiral {[}50{]} & 1.41 & 0.98 & 0.36 & 5.38 & 4.69 & \textbf{5.66}\tabularnewline
\hline 
D31 {[}50{]} & 2.33 & 1.35 & 1.21 & 4.40 & \textbf{5.36} & 3.35\tabularnewline
\hline 
Abalone {[}50{]} & 3.78 & 3.91 & 1.29 & 6.28 & \textbf{6.32} & 6.27\tabularnewline
\hline 
Car {[}50{]} & 1.76 & 1.76 & 1.57 & \textbf{3.02} & 3.00 & 2.97\tabularnewline
\hline 
Musk {[}50{]} & 2.27 & 2.83 & 0.01 & 2.69 & \textbf{3.87} & 3.64\tabularnewline
\hline 
Shuttle {[}100{]} & 1.86 & 1.84 & 1.32 & \textbf{4.04} & 3.66 & 3.71\tabularnewline
\hline 
\end{tabular}}
\par\end{centering}
\caption{The DB index of the clustering methods on the experimental datasets.
Larger is better. The number (i.e., {[}x{]}) nearby the dataset name
specifies the budget size used in the proposed method.\label{tab:com_db}}
\end{table}

\begin{table}[H]
\begin{centering}
\resizebox{0.95\textwidth}{!}{%
\begin{tabular}{|c|c|c|c|c|c|c|}
\hline 
\multirow{2}{*}{\textbf{\small{}Dataset}} & \multicolumn{6}{c|}{\textbf{Training Time}}\tabularnewline
\cline{2-7} 
 & \textbf{\small{}SVC} & \textbf{\small{}SGD} & \textbf{FSVC} & \textbf{BSGD-R} & \textbf{BSGD-PNN} & \textbf{BSGD-PRN}\tabularnewline
\hline 
Aggregation {[}50{]}  & 0.05 & 0.03 & 0.05 & \textbf{0.01} & 0.02 & 0.03\tabularnewline
\hline 
Breast Cancer {[}50{]} & 0.18 & \textbf{0.02} & 0.05 & \textbf{0.02} & 0.03 & 0.04\tabularnewline
\hline 
Compound {[}50{]} & 0.03 & 0.02 & 0.10 & \textbf{0.01} & \textbf{0.01} & \textbf{0.01}\tabularnewline
\hline 
Flame {[}50{]} & 0.02 & 0.02 & 15.16 & \textbf{0.01} & \textbf{0.01} & \textbf{0.01}\tabularnewline
\hline 
Glass {[}50{]} & 0.03 & 0.03 & 0.02 & \textbf{0.01} & \textbf{0.01} & \textbf{0.01}\tabularnewline
\hline 
Iris {[}50{]} & 0.02 & 0.02 & 0.04 & \textbf{0.01} & \textbf{0.01} & \textbf{0.01}\tabularnewline
\hline 
Jain {[}50{]} & 0.02 & 0.02 & 0.03 & \textbf{0.01} & \textbf{0.01} & \textbf{0.01}\tabularnewline
\hline 
Pathbased {[}50{]} & 0.02 & 0.02 & 0.05 & \textbf{0.01} & \textbf{0.01} & \textbf{0.01}\tabularnewline
\hline 
R15 {[}50{]} & 0.02 & 0.02 & 0.02 & \textbf{0.01} & \textbf{0.01} & 0.02\tabularnewline
\hline 
Spiral {[}50{]} & 0.02 & 0.03 & 0.02 & \textbf{0.01} & \textbf{0.01} & \textbf{0.01}\tabularnewline
\hline 
D31 {[}50{]} & 0.17 & \textbf{0.09} & \textbf{0.09} & 0.13 & 0.16 & 0.11\tabularnewline
\hline 
Abalone {[}50{]} & 2.26 & 0.81 & 10.94 & 0.36 & 0.27 & \textbf{0.22}\tabularnewline
\hline 
Car {[}50{]} & 5.62 & 0.64 & 8.15 & \textbf{0.05} & 0.06 & 0.07\tabularnewline
\hline 
Musk {[}50{]} & 55.93 & 5.79 & 58.49 & \textbf{1.36} & 1.89 & 1.43\tabularnewline
\hline 
Shuttle {[}100{]} & 10.03 & \textbf{0.46} & 68.43 & 1.51 & 2.15 & 1.82\tabularnewline
\hline 
\end{tabular}}
\par\end{centering}
\caption{The training time (in second) of the clustering methods on the experimental
datasets. Shorter is better. The number (i.e., {[}x{]}) nearby the
dataset name specifies the budget size used in the proposed method.\label{tab:runtime}}
\end{table}

\begin{table}[H]
\begin{centering}
\resizebox{0.95\textwidth}{!}{%
\begin{tabular}{|c|c|c|c|c|c|c|}
\hline 
\multirow{2}{*}{\textbf{\small{}Dataset}} & \multicolumn{6}{c|}{\textbf{Clustering Time}}\tabularnewline
\cline{2-7} 
 & \textbf{\small{}SVC} & \textbf{\small{}SGD} & \textbf{FSVC} & \textbf{BSGD-R} & \textbf{BSGD-PNN} & \textbf{BSGD-PRN}\tabularnewline
\hline 
Aggregation {[}50{]}  & 31.42 & 2.83 & 7.51 & \textbf{0.37} & 0.61 & 1.43\tabularnewline
\hline 
Breast Cancer {[}50{]} & 19.80 & 2.14 & 22.86 & \textbf{0.29} & 0.62 & 1.48\tabularnewline
\hline 
Compound {[}50{]} & 6.82 & 1.17 & 7.24 & \textbf{0.11} & 0.15 & 0.51\tabularnewline
\hline 
Flame {[}50{]} & 1.81 & 0.67 & 4.31 & \textbf{0.05} & 0.08 & 0.24\tabularnewline
\hline 
Glass {[}50{]} & 2.30 & 0.53 & 10.67 & \textbf{0.06} & 0.09 & 0.41\tabularnewline
\hline 
Iris {[}50{]} & 1.03 & 0.34 & 4.33 & \textbf{0.03} & \textbf{0.03} & 0.15\tabularnewline
\hline 
Jain {[}50{]} & 5.80 & 0.81 & 4.59 & \textbf{0.10} & 0.16 & 0.45\tabularnewline
\hline 
Pathbased {[}50{]} & 4.02 & 0.54 & 4.22 & \textbf{0.06} & 0.09 & 0.33\tabularnewline
\hline 
R15 {[}50{]} & 4.14 & 3.68 & 10.43 & \textbf{0.22} & 0.35 & 0.94\tabularnewline
\hline 
Spiral {[}50{]} & 1.60 & 0.99 & 7.78 & \textbf{0.07} & 0.10 & 0.35\tabularnewline
\hline 
D31 {[}50{]} & 467.72 & 6.56 & 33.08 & \textbf{4.58} & 8.39 & 13.33\tabularnewline
\hline 
Abalone {[}50{]} & 653.65 & 26.58 & 242.97 & 3.79 & \textbf{1.84} & 5.57\tabularnewline
\hline 
Car {[}50{]} & 67.66 & \textbf{7.05} & 84.47 & 16.24 & 8.42 & 23.57\tabularnewline
\hline 
Musk {[}50{]} & 602.09 & 432.58 & 510.25 & \textbf{173.36} & 194.69 & 183.78\tabularnewline
\hline 
Shuttle {[}100{]} & 1,972.61 & \textbf{925} & 1,125.46 & 1,766.66 & 1,739.62 & 1,759.06\tabularnewline
\hline 
\end{tabular}}
\par\end{centering}
\caption{The clustering time (in second) of the clustering methods on the experimental
datasets. Shorter is better. The number (i.e., {[}x{]}) nearby the
dataset name specifies the budget size used in the proposed method.\label{tab:clustime}}
\end{table}

We use $5$ cluster validation indices including purity, rand index,
NMI, DB index, and compactness to compare the proposed method with
the baselines. Each experiment is carried out $5$ times and the averages
of the cluster validation indices are reported. For readability, we
emphasize in boldface the method that yields the best CVI for each
dataset. Regarding purity (cf. Table \ref{tab:pu_rand_nmi}), our
proposed method outperforms other baselines. Its three variants obtain
the best purity on $12$ out of $15$ experimental datasets. This
fact also indicates a good classification ability of the proposed
method. Hence, the proposed method including its three variants can
be certainly used for classification purpose. As regards rand index
(cf. Table \ref{tab:rand}), the proposed method gains comparable
results in comparison to the baselines. In reference to NMI (cf. Table
\ref{tab:nmi}), the proposed method is again comparable with the
baselines. In particular, its three variants win over $10$ out of
$15$ experimental datasets. The proposed method including its three
variants also attains very convincing compactness indices comparing
with the baselines (cf. Table \ref{tab:com}). In particular, its
three variants surpass others on $11$ experimental datasets. The
last reported CVI is DB index. The proposed method dominates others
on this CVI (cf. Table \ref{tab:com_db}). In particular, it exceeds
others on all experimental datasets. Regarding the amount of time
taken for finding the domain of novelty, the proposed method especially
the variant BSGD-R is faster than the baselines on most datasets.
The proposed method and SGD-LMOC \cite{pham2016fast} are SGD-based
method but because of its sparser model, the proposed method is faster
than SGD-LMOC whilst preserving the learning performance of this method.
In regards to the cluster time, the proposed method again outperforms
the baselines. Comparing among three variants of the proposed method,
BSGD-R (i.e., removal strategy) is faster than others. It is reasonable
from the simplicity of this strategy. The BSGD-PNN variant (i.e.,
projection with $k$-nearest neighbors) is a little slower than the
BSGD-PRN (i.e., projection with $k$-random neighbors). The reason
lies in the fact that BSGD-PNN needs to sort out the support vectors
in the model which raises more computational burden.

\section{Conclusion}

In this paper, we have proposed a scalable support-based clustering
method, which conjoins the advantages of SGD-based method, kernel-based
method, and budget approach. Furthermore, we have also proposed a
new strategy for clustering assignment. We have validated our proposed
method on $15$ well-known datasets for clustering. The experimental
results have shown that our proposed method has achieved the comparable
clustering quality comparing with the baseline whilst being much faster.

\clearpage{}

\section*{References}

\bibliographystyle{plain}

\clearpage{}

\appendix

\section{Proofs regarding convergence analysis}

In this appendix, we display all proofs regarding convergence analysis
of BSGD-LMOC.

\textbf{Proof of Lemma \ref{lem:st}}
\begin{align*}
s_{k+1} & \leq\frac{k-1}{k}s_{k}+\left|\alpha_{n_{k}}\right|R\leq\frac{k-1}{k}s_{k}+\frac{CR}{k}\\
ks_{k+1} & \leq\left(k-1\right)s_{k}+CR
\end{align*}
Summing when $k=1,...,t-1$, we gain
\begin{align*}
\left(t-1\right)s_{t} & \leq\left(t-1\right)CR\\
s_{t} & \leq CR
\end{align*}

\textbf{Proof of Lemma \ref{lem:wt}}
\begin{align*}
\bw_{t} & =\sum_{i\in I_{t}}\alpha_{i}\phi\left(x_{i}\right)\\
\norm{\bw_{t}} & \leq\sum_{i\in I_{t}}\left|\alpha_{i}\right|R\leq Rs_{t}\leq CR^{2}
\end{align*}

\textbf{Proof of Lemma \ref{lem:gt}}
\begin{align*}
g_{t} & =\bw_{t}-C\mathbb{I}_{y_{n_{t}}\transp{\bw}\phi\left(x_{n_{t}}\right)<\theta_{nt}}y_{n_{t}}\phi\left(x_{n_{t}}\right)\\
\norm{g_{t}} & \leq\norm{\bw_{t}}+CR\leq CR^{2}+CR
\end{align*}

\textbf{Proof of Lemma \ref{lem:remove}}

We assume that the date sample $\phi\left(x_{p_{t}}\right)$ is sampled
at $m$ iterations $k_{1},k_{2},...,k_{m}$. At the iteration $k_{i}$,
$\phi\left(x_{p_{t}}\right)$ arrives in and the correspondent coefficient
is added by $\frac{-Cl^{'}\left(\bw_{k_{i}};x_{p_{t}},y_{p_{t}}\right)}{k_{i}}$
where $l\left(\bw;x,y\right)=\max\left\{ 0,\theta-y\transp{\bw}\phi\left(x\right)\right\} $
with $\theta=\left(1-y\right)/2$.

At the iteration $t$, the above quantity becomes
\begin{gather*}
\frac{t-1}{t}\times\frac{t-2}{t-1}\times...\times\frac{k_{i}}{k_{i}+1}\times\frac{-Cl^{'}\left(\bw_{k_{i}};x_{p_{t}},y_{p_{t}}\right)}{k_{i}}=\frac{-Cl^{'}\left(\bw_{k_{i}};x_{p_{t}},y_{p_{t}}\right)}{t}
\end{gather*}
Therefore, the following is guaranteed
\[
\left|\alpha_{p_{t}}\right|\leq\sum_{i=1}^{m}\norm{\frac{Cl^{'}\left(\bw_{k_{i}};x_{p_{t}},y_{p_{t}}\right)}{t}}\leq\frac{mCR}{t}
\]

\textbf{Proof of Lemma \ref{lem:ht}}
\[
\norm{h_{t}}\leq\left|\rho_{p_{t}}\right|\norm{\phi\left(x_{p_{t}}\right)}=t\left|\alpha_{p_{t}}\right|\norm{\phi\left(x_{p_{t}}\right)}\leq mCR^{2}=H
\]

\textbf{Proof of Lemma \ref{lem:wtw}}
\begin{flalign*}
\norm{\bw_{t+1}-\bw^{*}}^{2} & =\norm{\bw_{t}-\eta_{t}g_{t}-Z_{t}\alpha_{p_{t}}\phi\left(x_{p_{t}}\right)-\bw^{*}}^{2}=\norm{\bw_{t}-\eta_{t}g_{t}-\eta_{t}h_{t}-\bw^{*}}^{2}\\
 & =\norm{\bw_{t}-\bw^{*}}^{2}+\eta_{t}^{2}\norm{g_{t}+h_{t}}^{2}-2\eta_{t}\transp{\left(\bw_{t}-\bw^{*}\right)}g_{t}-2\eta_{t}\transp{\left(\bw_{t}-\bw^{*}\right)}h_{t}
\end{flalign*}
Taking the conditional expectation w.r.t $\bw_{t}$, we gain
\begin{flalign*}
\mathbb{E}\left[\norm{\bw_{t+1}-\bw^{*}}^{2}\right] & \leq\mathbb{E}\left[\norm{\bw_{t}-\bw^{*}}^{2}\right]+\eta_{t}^{2}\left(G+H\right)^{2}\\
 & -2\eta_{t}\transp{\left(\bw_{t}-\bw^{*}\right)}\mathbb{E}\left[g_{t}\right]-2\eta_{t}\transp{\left(\bw_{t}-\bw^{*}\right)}\mathbb{E}\left[h_{t}\right]\\
 & \leq\mathbb{E}\left[\norm{\bw_{t}-\bw^{*}}^{2}\right]+\eta_{t}^{2}\left(G+H\right)^{2}\\
 & -2\eta_{t}\transp{\left(\bw_{t}-\bw^{*}\right)}\mathcal{J}^{'}\left(\bw_{t}\right)-2\eta_{t}\transp{\left(\bw_{t}-\bw^{*}\right)}\mathbb{E}\left[h_{t}\right]\\
 & \leq\mathbb{E}\left[\norm{\bw_{t}-\bw^{*}}^{2}\right]+\eta_{t}^{2}\left(G+H\right)^{2}\\
 & -\frac{2\eta_{t}\norm{\bw_{t}-\bw^{*}}^{2}}{2}-2\eta_{t}\transp{\left(\bw_{t}-\bw^{*}\right)}\mathbb{E}\left[h_{t}\right]
\end{flalign*}
Here we note that we have used the following inequality
\begin{align*}
\transp{\left(\bw_{t}-\bw^{*}\right)}\mathcal{J}^{'}\left(\bw_{t}\right) & \geq\mathcal{J}\left(\bw_{t}\right)-\mathcal{J}\left(\bw^{*}\right)+\frac{1}{2}\norm{\bw_{t}-\bw^{*}}^{2}\geq\frac{1}{2}\norm{\bw_{t}-\bw^{*}}^{2}
\end{align*}

\noindent Taking the expectation again, we achieve
\begin{flalign*}
\mathbb{E}\left[\norm{\bw_{t+1}-\bw^{*}}^{2}\right] & \leq\frac{t-1}{t}\mathbb{E}\left[\norm{\bw_{t}-\bw^{*}}^{2}\right]+\eta_{t}^{2}\left(G+H\right)^{2}-2\eta_{t}\mathbb{E}\left[\transp{\left(\bw_{t}-\bw^{*}\right)}h_{t}\right]\\
 & \leq\frac{t-1}{t}\mathbb{E}\left[\norm{\bw_{t}-\bw^{*}}^{2}\right]+\eta_{t}^{2}\left(G+H\right)^{2}\\
 & +2\eta_{t}\mathbb{E}\left[\norm{\bw_{t}-\bw^{*}}^{2}\right]^{1/2}\mathbb{E}\left[\norm{h_{t}}^{2}\right]^{1/2}\\
 & \leq\frac{t-1}{t}\mathbb{E}\left[\norm{\bw_{t}-\bw^{*}}^{2}\right]+\frac{\left(G+H\right)^{2}}{t}+\frac{2H\mathbb{E}\left[\norm{\bw_{t}-\bw^{*}}^{2}\right]^{1/2}}{t}
\end{flalign*}
Choosing $W=H+\sqrt{H^{2}+\left(G+H\right)^{2}}$, we have the following:
if $\mathbb{E}\left[\norm{\bw_{t}-\bw^{*}}^{2}\right]\leq W^{2}$
, $\mathbb{E}\left[\norm{\bw_{t+1}-\bw^{*}}^{2}\right]\leq W^{2}$.

\textbf{Proof of Theorem \ref{thm:regret}}
\begin{flalign*}
\norm{\bw_{t+1}-\bw^{*}}^{2} & =\norm{\bw_{t}-\eta_{t}g_{t}-Z_{t}\alpha_{p_{t}}\phi\left(x_{p_{t}}\right)-\bw^{*}}^{2}=\norm{\bw_{t}-\eta_{t}g_{t}-\eta_{t}h_{t}-\bw^{*}}^{2}\\
 & =\norm{\bw_{t}-\bw^{*}}^{2}+\eta_{t}^{2}\norm{g_{t}+h_{t}}^{2}-2\eta_{t}\transp{\left(\bw_{t}-\bw^{*}\right)}g_{t}-2\eta_{t}\transp{\left(\bw_{t}-\bw^{*}\right)}h_{t}
\end{flalign*}
\begin{flalign*}
\transp{\left(\bw_{t}-\bw^{*}\right)}g_{t} & =\frac{\norm{\bw_{t}-\bw^{*}}^{2}-\norm{\bw_{t+1}-\bw^{*}}^{2}}{2\eta_{t}}+\frac{\eta_{t}\norm{g_{t}+h_{t}}^{2}}{2}-\transp{\left(\bw_{t}-\bw^{*}\right)}h_{t}\\
 & =\frac{\norm{\bw_{t}-\bw^{*}}^{2}-\norm{\bw_{t+1}-\bw^{*}}^{2}}{2\eta_{t}}+\frac{\eta_{t}\norm{g_{t}+h_{t}}^{2}}{2}-\transp{\left(\bw_{t}-\bw^{*}\right)}Z_{t}\rho_{p_{t}}\phi\left(x_{p_{t}}\right)
\end{flalign*}
Taking the conditional expectation w.r.t $\bw_{1},...,\bw_{t},\,x_{1},\ldots,x_{t}$
of two sides on the above inequality, we gain
\begin{flalign*}
\transp{\left(\bw_{t}-\bw^{*}\right)}\mathcal{J}^{'}\left(\bw_{t}\right) & \leq\frac{\mathbb{E}\left[\norm{\bw_{t}-\bw^{*}}^{2}\right]-\mathbb{E}\left[\norm{\bw_{t+1}-\bw^{*}}^{2}\right]}{2\eta_{t}}\\
 & +\frac{\eta_{t}\left(G+H\right)^{2}}{2}-\transp{\left(\bw_{t}-\bw^{*}\right)}\mathbb{E}\left[Z_{t}\right]\rho_{p_{t}}\phi\left(x_{p_{t}}\right)
\end{flalign*}
\begin{flalign*}
\mathcal{J}\left(\bw_{t}\right)-\mathcal{J}\left(\bw^{*}\right)+\frac{1}{2}\norm{\bw_{t}-\bw^{*}}^{2} & \leq\frac{\mathbb{E}\left[\norm{\bw_{t}-\bw^{*}}^{2}\right]-\mathbb{E}\left[\norm{\bw_{t+1}-\bw^{*}}^{2}\right]}{2\eta_{t}}\\
 & +\frac{\eta_{t}\left(G+H\right)^{2}}{2}-\mathbb{P}\left(Z_{t}=1\right)\rho_{p_{t}}\transp{\left(\bw_{t}-\bw^{*}\right)}\phi\left(x_{p_{t}}\right)
\end{flalign*}
Here we note that $\rho_{p_{t}}$ is functionally dependent on $\bw_{1},...,\bw_{t},\,x_{1},\ldots,x_{t}$
and $\mathcal{J}\left(\bw\right)$ is 1-strongly convex function.

\noindent Taking the expectation of two sides of the above inequality,
we obtain
\begin{flalign*}
\mathbb{E}\left[\mathcal{J}\left(\bw_{t}\right)-\mathcal{J}\left(\bw^{*}\right)\right] & \leq\frac{t-1}{2}\mathbb{E}\left[\norm{\bw_{t}-\bw^{*}}^{2}\right]-\frac{t}{2}\mathbb{E}\left[\norm{\bw_{t+1}-\bw^{*}}^{2}\right]\\
 & +\frac{\left(G+H\right)^{2}}{2t}-\mathbb{P}\left(Z_{t}=1\right)\mathbb{E}\left[\transp{\left(\bw_{t}-\bw^{*}\right)}\rho_{p_{t}}\phi\left(x_{p_{t}}\right)\right]
\end{flalign*}
\begin{flalign*}
\mathbb{E}\left[\mathcal{J}\left(\bw_{t}\right)-\mathcal{J}\left(\bw^{*}\right)\right] & \leq\frac{t-1}{2}\mathbb{E}\left[\norm{\bw_{t}-\bw^{*}}^{2}\right]-\frac{t}{2}\mathbb{E}\left[\norm{\bw_{t+1}-\bw^{*}}^{2}\right]+\frac{\left(G+H\right)^{2}}{2t}\\
 & +\mathbb{P}\left(Z_{t}=1\right)\mathbb{E}\left[\norm{\bw_{t}-\bw^{*}}^{2}\right]^{1/2}\mathbb{E}\left[\norm{\rho_{p_{t}}\phi\left(x_{p_{t}}\right)}^{2}\right]^{1/2}\\
 & \leq\frac{t-1}{2}\mathbb{E}\left[\norm{\bw_{t}-\bw^{*}}^{2}\right]-\frac{t}{2}\mathbb{E}\left[\norm{\bw_{t+1}-\bw^{*}}^{2}\right]\\
 & +\frac{\left(G+H\right)^{2}}{2t}+\mathbb{P}\left(Z_{t}=1\right)RW\mathbb{E}\left[\rho_{p_{t}}^{2}\right]^{1/2}
\end{flalign*}
Taking sum when $t=1,...,T$, we gain
\begin{gather*}
\sum_{t=1}^{T}\mathbb{E}\left[\mathcal{J}\left(\bw_{t}\right)-\mathcal{J}\left(\bw^{*}\right)\right]\leq\frac{\left(G+H\right)^{2}}{2}\sum_{t=1}^{T}\frac{1}{t}+RW\sum_{t=1}^{T}\mathbb{P}\left(Z_{t}=1\right)\mathbb{E}\left[\rho_{p_{t}}^{2}\right]^{1/2}
\end{gather*}
\begin{gather*}
\frac{1}{T}\sum_{t=1}^{T}\mathbb{E}\left[\mathcal{J}\left(\bw_{t}\right)\right]-\mathcal{J}\left(\bw^{*}\right)\leq\frac{\left(G+H\right)^{2}\left(\log T+1\right)}{2T}+\frac{WR}{T}\sum_{t=1}^{T}\mathbb{P}\left(Z_{t}=1\right)\mathbb{E}\left[\rho_{p_{t}}^{2}\right]^{1/2}
\end{gather*}
\begin{gather*}
\mathbb{E}\left[\mathcal{J}\left(\overline{\bw}_{t}\right)\right]-\mathcal{J}\left(\bw^{*}\right)\leq\frac{1}{T}\sum_{t=1}^{T}\mathbb{E}\left[\mathcal{J}\left(\bw_{t}\right)\right]-\mathcal{J}\left(\bw^{*}\right)\leq\frac{\left(G+H\right)^{2}\left(\log T+1\right)}{2T}\\
+\frac{WR}{T}\sum_{t=1}^{T}\mathbb{P}\left(Z_{t}=1\right)\mathbb{E}\left[\rho_{p_{t}}^{2}\right]^{1/2}
\end{gather*}

\end{document}